\documentclass{article}
\usepackage{spconf,amsmath,graphicx,hyperref}
\usepackage{amssymb}
\usepackage{pifont}
\newcommand{\cmark}{\ding{51}}%
\newcommand{\xmark}{\ding{55}}
\usepackage{multirow}
\usepackage{makecell}

\usepackage{xcolor}

\title{Multiview Progress Prediction of Robot Activities}
\name{Elena Zoppellari$^{\star}$ \qquad Federico Becattini$^{\dagger}$ \qquad Marco Fiorucci$^{\star}$ \qquad Lamberto Ballan$^{\star}$}
  
\address{$^{\star}$ University of Padova, Italy \qquad
      $^{\dagger}$ University of Siena, Italy}
\begin{document}
%
\maketitle
\begin{abstract}
For robots to operate effectively and safely alongside humans, they must be able to understand the progress of ongoing actions. This ability, known as action progress prediction, is critical for tasks ranging from timely assistance to autonomous decision-making. However, modeling action progression in robotics has often been overlooked. Moreover, a single camera may be insufficient for understanding robot's ego-actions, as self-occlusion can significantly hinder perception and model performance. In this paper, we propose a multi-view architecture for action progress prediction in robot manipulation tasks. Experiments on Mobile ALOHA demonstrate the effectiveness of the proposed approach.
\end{abstract}
\section{Introduction}
\label{sec:intro}
Robots are increasingly entering human-centric environments, from factories to homes. To act as intelligent partners rather than mere tools, they must accurately understand actions performed by humans, other robots, or themselves. A fundamental, but often overlooked, aspect of this understanding is the concept of action progress: knowing not just what action is being performed, but how far along it is towards completion. The ability to estimate action progress is fundamental for safe and efficient human-robot interaction, enabling robots to anticipate human needs, provide timely assistance, and seamlessly coordinate joint tasks. It also benefits autonomous operation in non-collaborative environments by allowing robots to detect mistakes and adjust their maneuvers accordingly. 

The task of estimating action progress was introduced by Becattini et al. \cite{becattini2020done} for spatio-temporal action localization and has then been adopted in several contexts, spanning from human activity understanding \cite{qu2020lap} to tracking surgical procedures \cite{wang2023real}, but remains largely unexplored in robotics.
Furthermore, research in the field has focused on analyzing actions from a single viewpoint, which limits applicability in robotic settings. The robot's own arms may occlude the workspace, or a complex manipulation task might require viewpoints from multiple angles to be fully understood, making single-camera setups insufficient for robust action understanding.

To address these limitations, we introduce a multi-view approach for action progress prediction, tailored specifically for robotic manipulation.
Our central hypothesis is that by fusing information from multiple synchronized cameras, a model can gain a more comprehensive and robust representation of the action, overcoming occlusions and capturing spatial relationships that are invisible from a single perspective. We utilize a setup with three cameras mounted on a robot's head and arms, which reflects a practical configuration for modern mobile manipulators.

In summary, the main contributions of this work are the following: i) we highlight the significance of action progress prediction as a core capability for intelligent robotic systems; ii) we propose a novel multi-stream deep learning architecture that effectively fuses information from multiple camera views to estimate action progress; iii) our results demonstrate that our approach significantly outperforms single-view methods. We also analyze the contribution of each camera and we present different training strategies necessary to avoid optimization shortcuts and trivial solutions.

\section{Related Works}
\label{sec:related}

A few early works have studied tasks related to action progress, finalized either to early action detection \cite{ma2016learning} or phase detection and action completion \cite{li2017progress}.
The work of Becattini et al. \cite{becattini2020done} formally defined the action progress prediction task, distinguishing it from related tasks such as action recognition or detection. They proposed a unified view rooted in linguistics, along with an initial methodology and a model called ProgressNet, which combines a CNN for spatial features with an LSTM to model temporal progress. 
Building on this work, subsequent research has advanced the field by integrating progress prediction with action-related tasks, such as temporal action detection \cite{lu2024action}.
While progress is typically modeled as a continuous function, some studies have explored discrete formulations for online action detection \cite{qu2020lap}, for instance, using granularities ranging from 5 to 20 bins~\cite{han2017human}.
Progress prediction has also been formulated as the estimation of the remaining duration, particularly in the surgical domain~\cite{aksamentov2017deep, marafioti2021catanet, twinanda2018rsdnet, wang2023real}.

A different take on action progress was proposed by Donahue et al \cite{donahue2024learning}, who introduced a self-supervised paradigm based on video alignment to learn activity progression.
Similarly, Dwibedi et al. \cite{dwibedi2019temporal} explored self-supervised video alignment by leveraging cycle consistency between frames from different videos. Their method enabled the learning of action-specific feature extractors and, as a byproduct, demonstrated the model's ability to capture activity progression.

Action progress estimation has extended beyond RGB inputs. Hu et al. \cite{hu2019progress} used a combination of RGB and optical flow for spatio-temporal action detection, using progress to guide the tube generation step by selecting boxes for which progress increases.
Additionally, action progress estimation directly from 3D human joints has proven to be extremely effective~\cite{pucci2023joint}.
Recently, De Boer et al. \cite{de2023there} raised a provocative research question, asking whether there is indeed progress in activity progress, stating that progress estimators are likely to exploit shortcuts without learning meaningful visual cues. We confirm this issue, but show it can be easily circumvented by well-designed data augmentation strategies.
In general, all works addressing action progress focus on single-view cameras, with no real application in robotics, with the exception of surgical robots. We address the problem of action progress from a robot's perspective, using a multi-camera setting exploiting three cameras placed on the head and the hands.

\section{Method}
\label{sec:method}
Let $\mathcal{S}$ be a stream composed of multiple views $\{V_k\}_{i=k,..,K}$, 
that yield synchronized frames $\{f^k_i\}_{i=1,...,N}$.
We formulate the task of multi-view action progress prediction as the one of learning a mapping $\Phi(\mathcal{S_{:\textit{t}}}) \rightarrow \hat{p}_t$, that approximates with $\hat{p}_t$ the true progress of the ongoing action $p_t$, observing all the frames up to the current time $t$.
Therefore, during the execution of the action, for each timestep $t$, the predictor parameterizing $\Phi$ can observe the current frame and all the past frames, and must predict the current progress, without accessing knowledge on future timestamps.
In this paper, we have $K=3$ concurrent views, corresponding to cameras mounted on the head of a robot (the $\texttt{central}$ camera) and on its two arms (the $\texttt{left}$ and $\texttt{right}$ cameras).

Following prior work \cite{becattini2020done, pucci2023joint}, we use a linear formulation of progress, assigning to each frame a progress value $p_t=t/(t_E - t_S)$, with $t_S$ and $t_E$ the starting and ending timestamp for the current action.
The architecture we propose for estimating action progress is a multi-stream pipeline, composed of a visual backbone, dedicated to feature extraction from the three cameras of the robot, and a temporal model, accumulating knowledge of past frames in its internal state.
The backbone, in principle, can be embodied by any frame-based feature extractor. In our experiments, we leverage the vision transformer ViT-B/16 \cite{dosovitskiy2020vit}, but we also test other backbones such as MobileNetV2 \cite{sandler2018mobilenetv2}, ResNet18 and ResNet152 \cite{he2016deep}.
In the ViT backbone, the vision transformer is used as a feature extractor by removing its classification head. The input images are first processed using a convolutional projection layer that produces patch embeddings. A learnable class token is concatenated to the sequence, and the resulting tokens are passed through the transformer encoder. The output tokens (excluding the class token) are reshaped back into a spatial layout, resulting in a grid of patch-wise feature vectors.
Instead, for the other backbones, we simply remove the classification head.

Since different backbones extract features with different shapes, we pass the feature extracted from the visual backbone through a Spatial Pyramid Pooling (SPP) module \cite{he2015spatial}. We use a pyramid depth of 3, meaning that three different versions of the feature are pooled, with spatial resolutions of $1\times1$, $2\times2$ and $3\times3$. The resulting feature is then processed by a fully connected layer with output shape $512$, followed by dropout and ReLU activation.
Finally, the features extracted by each view are concatenated into a single vector and processed by another sequence of fully connected (output shape $64$), dropout and ReLU, before being fed to two stacked LSTM layers with hidden size $32$. The LSTM thus predicts the progress for each frame in a causal way, i.e., predicting online as the current frame is observed.

\begin{figure}[t]
    \centering
    \includegraphics[width=.9\linewidth]{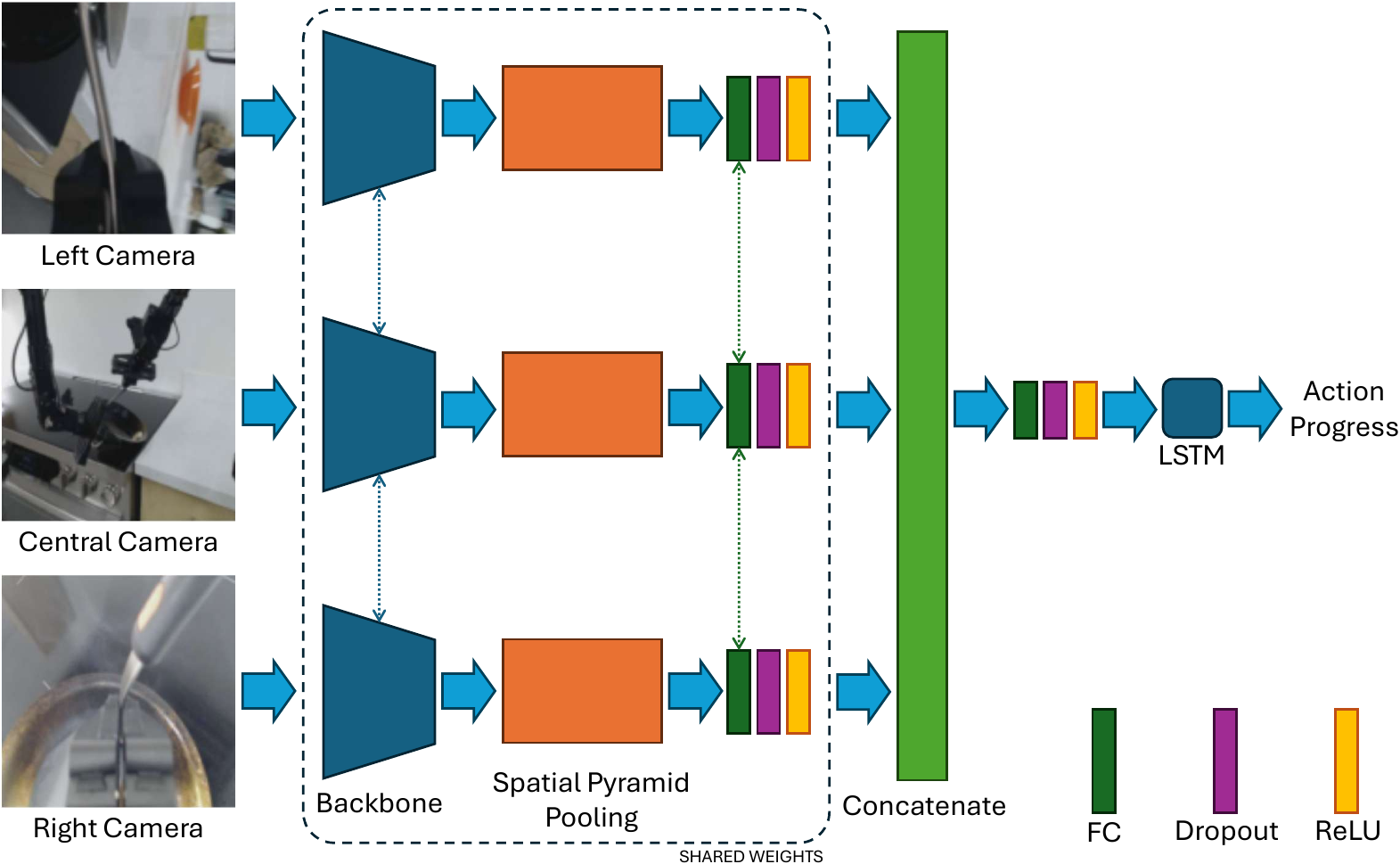}
    \caption{Proposed architecture}
    \label{fig:placeholder}
\end{figure}

\noindent\textbf{Training.} 
For training, we minimize a Mean Absolute Error (MAE) as loss using Adam, with learning rate $10^{-4}$, $\beta 1=0.9$ and $\beta 2=0.999$. The dropout rate was set to 0.5.
To avoid trivial solutions, such as constant predictions or solutions based on frame counting (issues observed in prior works \cite{becattini2020done, de2023there}), we introduce a few augmentations to encourage the model to learn from visual cues.
First, we introduce at training time a variable frame rate preprocessing to introduce temporal variability: randomly selecting different frame rates for different segments of a video, prevents the model from associating a specific duration with an action. Without this augmentation, the training often collapses to non-informative solutions.
Second, we exploit a segment-based training, which involves using randomly sampled portions of a video rather than the full, complete sequence. This technique is used to force a model to rely on visual cues for a task, instead of learning the temporal structure or duration of the entire action. By training on segments of varying lengths and starting points, the model cannot simply learn that the progress is a function of the frame index within the video.

\section{Experiments}
\label{sec:exp}

\noindent\textbf{Dataset}
\begin{table}[t]
\centering
\resizebox{.99\columnwidth}{!}{
\begin{tabular}{l|ll}
\textbf{Action} & \textbf{Start Condition} & \textbf{End Condition} \\
\hline
Use Cabinet & Base velocity $>$ 0.2 m/s & Shoulder or elbow velocity $<$ 0.4 m/s \\
Push Chair & Base angular velocity $>$ 0.3 rad/s & Either gripper effort $<$ -500 Nm \\
Take Elevator & Base velocity $>$ 0.3 m/s & Base angular velocity $<$ 0.3 rad/s \\
Cook Shrimp & Right gripper effort $>$ 200 Nm & Left gripper effort $<$ 600 Nm \\
Wash Pan & Left gripper effort $>$ 150 Nm & Left gripper effort $<$ 100 Nm \\
Wipe Wine & Base velocity $>$ 0.25 m/s & Shoulder, elbow or wrist angle vel. $<$ 0.5 m/s \\
\end{tabular}
}
\caption{Heuristic start and end conditions for each action in \textit{Mobile ALOHA} \cite{fu2024mobile}, based on robot motion signals.}
\label{tab:boundary_heuristics}
\end{table}
For our experiments, we use \textit{Mobile ALOHA}~\cite{fu2024mobile}, a recent dataset created for imitation learning with bimanual robots. It includes around 300 demonstrations, 6 distinct manipulation tasks and three synchronized camera views: one egocentric (head-mounted) and two external (arm-mounted).
Each frame includes detailed motion data from the robot’s base and arms. The base data provides linear and angular velocity, while each arm has seven degrees of freedom (DoF), with joint-level readings on position, velocity and effort, covering the waist, shoulder, elbow, forearm, wrist (angle and rotation) and gripper.
We preprocess each video sequence in the dataset to remove the initial and final parts of the videos, where the robot waits idly.
We manually defined a set of action-specific heuristic conditions to identify task boundaries by inspecting sensor traces.
Table~\ref{tab:boundary_heuristics} summarizes the resulting rules. These rules proved to be generally reliable, with the exception of a few edge cases in the \textit{use cabinet} sequences, which required manual adjustments.

\noindent\textbf{Baselines}
\label{sec:baselines}
We consider three baselines following prior work~\cite{becattini2020done, de2023there}: \texttt{Random predictions}: predicted progress is randomly sampled, uniformly between 0 and 1; \texttt{Static}: the model always predicts a fixed progress value of 0.5; \texttt{Average Index}: the predicted progress is computed as  $\hat{p}^i_n = \frac{1}{N_i} \sum_{m=1}^{N_i} p_m^i$, where $i$ is the frame index, $N_i$ is the number of training videos with at least $i$ frames, $m \in \{1, ..., N_i\}$ and $p_m^i$ is the ground-truth progress value for video $m$ at frame $i$.
These baselines provide reference points for evaluating the model’s ability to extract meaningful visual information. In fact, when quantitatively evaluated, the baselines can achieve surprisingly low errors and can help to highlight trivial solutions~\cite{becattini2020done}.

\begin{table}[t]
    \centering
    \resizebox{.99\columnwidth}{!}{
    \begin{tabular}{l|c|cccc|c}
        \textbf{Model} & \textbf{Segments} & \textbf{[0-25]\%} & \textbf{[25-50]\%} & \textbf{[50-75]\%} & \textbf{[75-100]\%} & \textbf{Whole}\\ \hline

        Avg. Index & - & 4.70 & 22.26 & 46.79 & 71.69 & 10.00\\
        Static & - & 37.46 & 12.51 & 12.44 & 37.39 & 25.01\\
        Random & - & 39.69 & 27.15 & 26.88 & 39.45 & 33.87 \\ \hline
        \multirow{2}{*}{MobileNet} & \xmark & 6.76 & 12.10 & 8.21 & 10.30 & 7.02\\
         & \cmark & 7.60 & 4.69 & 7.90 & 19.20 & 6.34\\ \hline

         \multirow{2}{*}{ResNet152} & \xmark & \textbf{3.24} & 5.30 & 31.90 & 61.00 & 6.57\\
         & \cmark & 7.60 & 4.69 & 7.90 & 19.2 & 6.57 \\ \hline

         \multirow{2}{*}{ResNet18} & \xmark & 4.46 & 5.93 & 6.64 & 10.40 & 5.35\\
         & \cmark & 5.25 & 4.75 & 6.94 & 19.20 & 5.32 \\ \hline

        \multirow{2}{*}{ViT} & \xmark & 4.00 & 4.52 & 5.64 & 6.82 & \textbf{4.86}\\
         & \cmark & 9.05 & \textbf{4.40} & \textbf{4.44} & \textbf{4.83} & 5.47 \\         
    \end{tabular}
    }
    \caption{MAE on Mobile ALOHA, evaluated on different progress intervals. Overall best result per progress interval highlighted in \textbf{bold}.}
    \label{tab:intervals}
\end{table}

\begin{table}[t]
    \centering
    \resizebox{.99\columnwidth}{!}{
    \begin{tabular}{l|c|cccc|c}
        \textbf{Camera} & \textbf{Segments} & \textbf{[0-25]\%} & \textbf{[25-50]\%} & \textbf{[50-75]\%} & \textbf{[75-100]\%} & \textbf{Whole} \\ \hline

        \multirow{2}{*}{Left} & \xmark & 8.58 & 11.20 & 14.90 & 30.70 & 9.98\\
         & \cmark & 20.80 & 6.63 & 5.47 & 9.27 & 10.10 \\ \hline

         \multirow{2}{*}{Right} & \xmark & 5.84 & 12.70 & 34.70 & 53.10 & 9.41 \\
         & \cmark & 24.50 & 15.70 & 9.09 & 9.75 & 15.60 \\ \hline

         \multirow{2}{*}{Central} & \xmark & 4.99 & 8.43 & 6.61 & 28.40 & 6.23\\
         & \cmark & 14.90 & 5.98 & 5.35 & 6.42 & 7.90\\ \hline

        \multirow{2}{*}{All} & \xmark & \textbf{4.00} & 4.52 & 5.64 & 6.82 & \textbf{4.86}\\
         & \cmark & 9.05 & \textbf{4.40} & \textbf{4.44} & \textbf{4.83} & 5.47 \\         
    \end{tabular}
    }
    \caption{MAE on Mobile ALOHA, evaluated on partial segments for different cameras with the ViT-based model. Overall best result per progress interval highlighted in \textbf{bold}.}
    \label{tab:camera_intervals}
\end{table}

\begin{table}[t]
    \centering
    \resizebox{.99\columnwidth}{!}{
    \begin{tabular}{l|c|cccccc}
          \textbf{Camera} & \textbf{Segments} & \textbf{\makecell{Use\\Cabinet}} & \textbf{\makecell{Push\\Chair}} & \textbf{\makecell{Take\\Elevator}} & \textbf{\makecell{Cook\\Shrimp}} & \textbf{\makecell{Wash\\Pan}} & \textbf{\makecell{Wipe\\Wine}} \\ \hline
         Left & \xmark & 8.48 & 13.00 & 8.18 & 12.80 & 8.49 & 7.57 \\
         Left & \cmark & 8.02 & 15.50 & 8.59 & 9.66 & 8.66 & 8.26 \\ \hline
         
         Right & \xmark & 8.36 & 10.30 & 13.40 & 8.05 & 10.60 & 7.55\\
         Right & \cmark & 12.40 & 21.20 & 26.80 & 11.80 & 12.00 & 12.10\\ \hline
         
         Central & \xmark & 5.90 & 5.37 & 5.03 & 6.85 & 7.55 & 7.27 \\
         Central & \cmark & 7.85 & 9.90 & 7.09 & 7.33 & 7.06 & 6.60 \\ \hline
         
         All & \xmark & \textbf{4.11} & \textbf{6.29} & \textbf{4.40} & \textbf{4.86} & 4.89 & 4.22 \\
         All & \cmark & 4.26 & 7.02 & 5.71 & 6.93 & \textbf{4.79} & \textbf{4.18} \\
    \end{tabular}
    }
    \caption{MAE per action on Mobile ALOHA (ViT+LSTM) using different cameras. Overall best result in \textbf{bold}.}
    \label{tab:cameras}
\end{table}

\begin{figure*}
    \centering
    \includegraphics[width=0.49\linewidth]{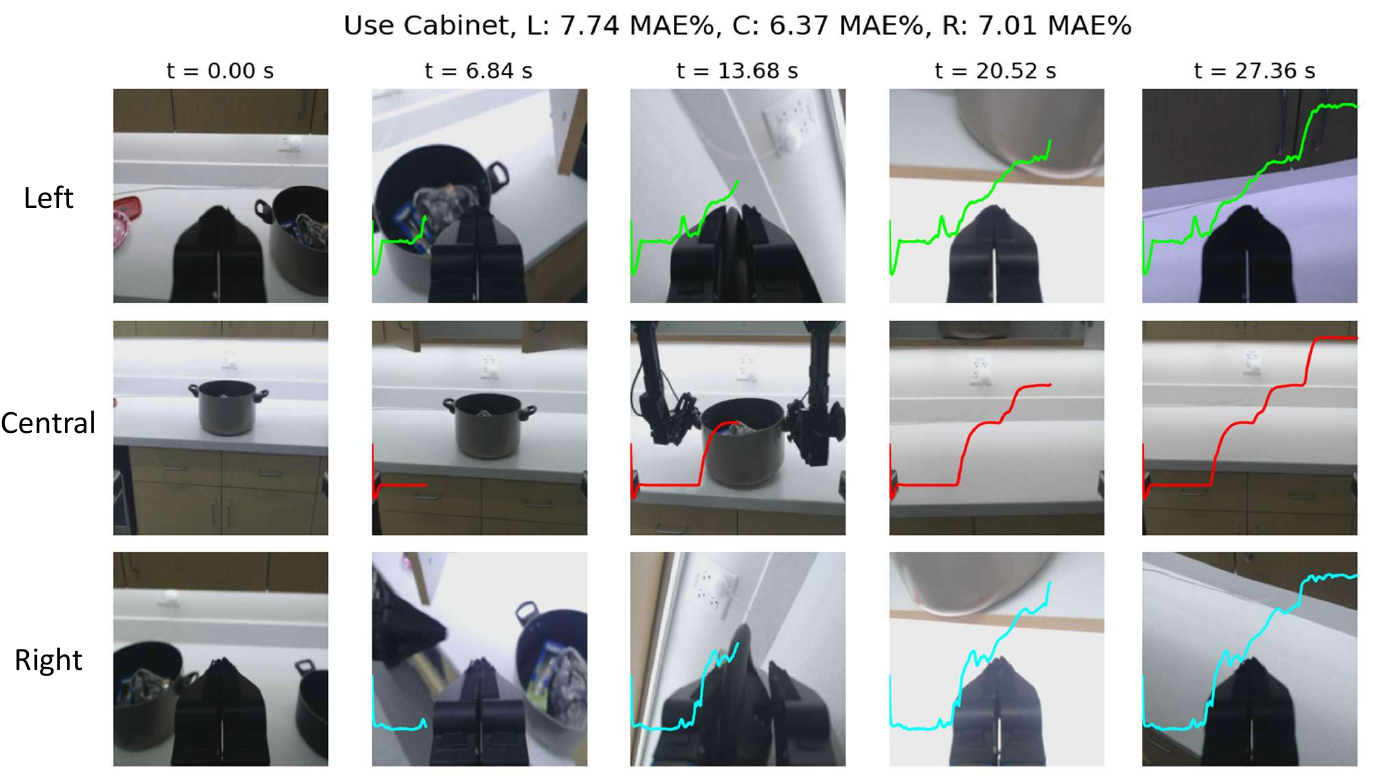} 
    \includegraphics[width=0.49\linewidth]{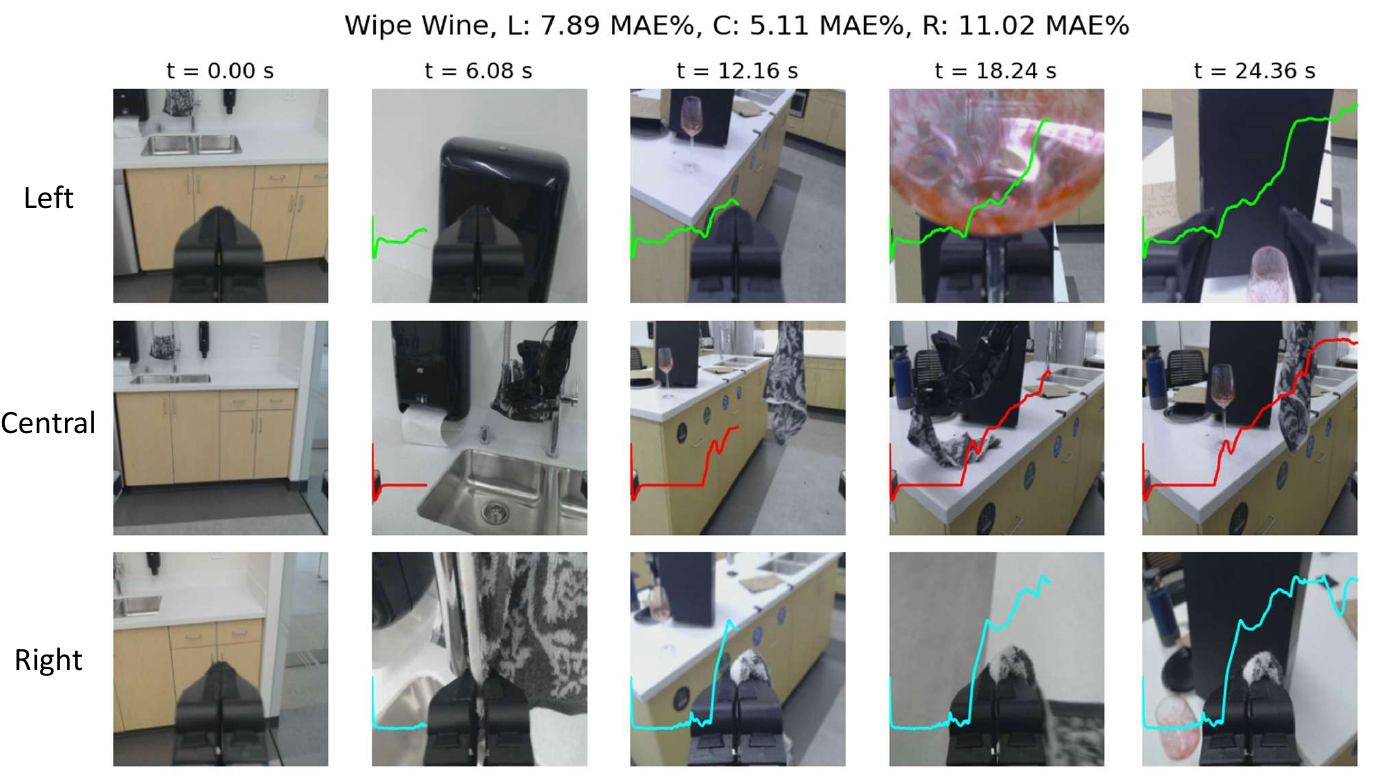} 
    \caption{Progress estimated from different cameras: left (green), central (red) and right (cyan).  A perfect progress estimation would exhibit a straight diagonal line from bottom left to top right.}
    \label{fig:qualitative_cameras}
\end{figure*}

\begin{figure*}
    \centering
    \includegraphics[width=0.49\linewidth]{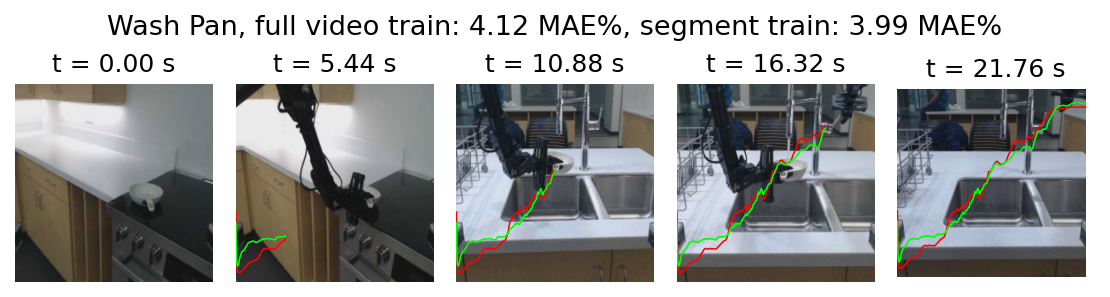} 
    \includegraphics[width=0.49\linewidth]{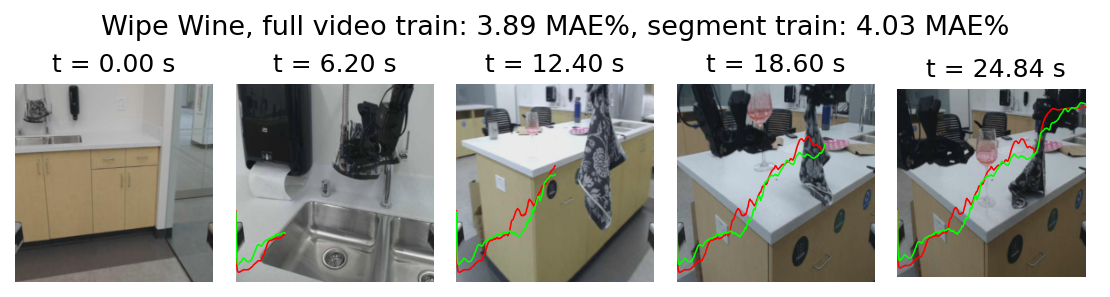} 
    \caption{Progress estimated with multiple cameras. Two versions are shown: trained on full videos (red) and trained on semgments (green).  A perfect progress estimation would exhibit a straight diagonal line from bottom left to top right.}
    \label{fig:qualitative_segments}
\end{figure*}

\noindent\textbf{Results}
\label{sec:results}
We report the main quantitative results in Tab. \ref{tab:intervals}, where we evaluate our approach with different visual backbones against several baselines using Mean Absolute Error (MAE) as the primary metric.
We report the results for the model evaluated on whole sequences, i.e., when the action is observed from the beginning to the end, as well as a breakdown into four quartiles according to ground truth progress. In this setting, each quartile is considered as a separate video. This setting is particularly important, as to shown if the model can handle partially observed videos. 
All our proposed models substantially outperform the Random, 0.5 Static Prediction, and Average Index baselines, which confirms that the network is learning meaningful visual features for progress estimation rather than adopting trivial solutions.

Of critical importance is our segment-based training strategy. Models trained on full sequences often fail to generalize, exhibiting extremely high error in the later portions of the action. This is particularly clear with the ResNet152 backbone, whose MAE reaches 61.00\% in the final quartile ([75-100]\%) of the sequence when trained without segments.
Among the different backbones trained with segments, the Vision Transformer (ViT) model demonstrates the most robust and stable performance. While a ResNet18-based model achieves a slightly lower overall MAE for the whole sequence (5.32\%), the ViT model is the top performer in the latter three progress intervals, achieving the lowest error in the [25-50]\%, [50-75]\%, and [75-100]\% ranges. This suggests that the ViT backbone is more effective at making accurate predictions as an action unfolds, especially as it nears completion.

We then performed an ablation study to evaluate the contribution of each camera view, using our best ViT-based model. The results are detailed in Tab. \ref{tab:camera_intervals} (per progress interval) and Tab. \ref{tab:cameras} (per action). 
The analysis of single-view performance reveals that the central camera, providing an egocentric perspective from the robot's head, is the most informative single source. It consistently achieves lower error rates compared to the arm-mounted cameras across most tasks and progress intervals. In fact, it is likely that the cameras on the arms only have a partial view or might be subject to strong occulusions, for example when manipulating objects.

Interestingly, the model that fuses information from all three cameras achieves a significant performance improvement, outperforming any single-view configuration. This multi-view approach obtains the best results on every individual action listed in Tab. \ref{tab:cameras}. For example, for the "Use Cabinet" task, fusing all views reduces the MAE to 4.11\%, a substantial improvement over the 5.90\% achieved by the best single camera (central). This demonstrates that integrating multiple perspectives provides a more comprehensive and robust representation for modeling action progress.


Finally, Fig. \ref{fig:qualitative_cameras} and Fig. \ref{fig:qualitative_segments} show qualitative results for our model with the ViT backbone.
The plots in Fig. \ref{fig:qualitative_cameras} visualize the predicted progress over time, with separate lines for the left (green), central (red), and right (cyan) cameras. These visualizations align with the quantitative data, often showing the central camera's prediction approximating the ground truth more closely.
Fig. \ref{fig:qualitative_segments} illustrates the impact of the segment-based training strategy. It compares the final multi-view prediction of a model trained on full videos (red line) against one trained on segments (green line). It can be seen that segments-based might yield slightly worse results in terms of MAE. However, this training makes the model react to visual cues, instead of simply counting frames or follow biases in the data, making the segments-based training more suited to comprehend the observed scene.

\section{Conclusions}
\label{sec:conclusions}

We presented a multi-view architecture for action progress prediction in robotics. Our experiments on the Mobile ALOHA dataset show that fusing information from head and arm-mounted cameras significantly improves prediction accuracy compared to any single view. While ablation studies identified the central egocentric camera as the most informative single source, the multi-view fusion consistently achieved superior performance by creating a more comprehensive spatio-temporal representation. We also demonstrated that a segment-based training strategy is critical for model generalization and avoiding trivial, non-visual solutions.

\textbf{Acknowledgments.} This work was partially supported by CARES (Climate-Aware Robotics for Energy Systems) PR FESR Toscana 2021-2027 CUP ST 17200.24072024.0630 00033, FEATHER (Forecasting and Estimation of Actions and Trajectories for Human–robot intERactions) Piano per lo Sviluppo della Ricerca PSR 2023 University of Siena, and the HPC CONVECS PR FESR Veneto
2021-2027 project.

\bibliographystyle{IEEEbib}
\bibliography{strings,refs}

@article{becattini2020done,
  title={Am i done? predicting action progress in videos},
  author={Becattini, Federico and Uricchio, Tiberio and Seidenari, Lorenzo and Ballan, Lamberto and Bimbo, Alberto Del},
  journal={ACM Transactions on Multimedia Computing, Communications, and Applications (TOMM)},
  volume={16},
  number={4},
  pages={1--24},
  year={2020},
  publisher={ACM New York, NY, USA}
}

@article{pucci2023joint,
  title={Joint-based action progress prediction},
  author={Pucci, Davide and Becattini, Federico and Del Bimbo, Alberto},
  journal={Sensors},
  volume={23},
  number={1},
  pages={520},
  year={2023},
  publisher={MDPI}
}

@inproceedings{donahue2024learning,
  title={Learning to predict activity progress by self-supervised video alignment},
  author={Donahue, Gerard and Elhamifar, Ehsan},
  booktitle={IEEE/CVF Conference on Computer Vision and Pattern Recognition},
  pages={18667--18677},
  year={2024}
}

@inproceedings{de2023there,
  title={Is there progress in activity progress prediction?},
  author={De Boer, Frans and van Gemert, Jan C and Dijkstra, Jouke and Pintea, Silvia L},
  booktitle={IEEE/CVF Conference on Computer Vision and Pattern Recognition},
  pages={2958--2966},
  year={2023}
}

@article{hu2019progress,
  title={Progress regression RNN for online spatial-temporal action localization in unconstrained videos},
  author={Hu, Bo and Cai, Jianfei and Cham, Tat-Jen and Yuan, Junsong},
  journal={arXiv preprint arXiv:1903.00304},
  year={2019}
}

@article{lu2024action,
  title={Action progression networks for temporal action detection in videos},
  author={Lu, Chong-Kai and Mak, Man-Wai and Li, Ruimin and Chi, Zheru and Fu, Hong},
  journal={IEEE Access},
  year={2024},
  publisher={IEEE}
}

@inproceedings{ma2016learning,
  title={Learning activity progression in lstms for activity detection and early detection},
  author={Ma, Shugao and Sigal, Leonid and Sclaroff, Stan},
  booktitle={IEEE Int'l Conference on Computer vision and Pattern Recognition},
  pages={1942--1950},
  year={2016}
}

@article{li2017progress,
  title={Progress estimation and phase detection for sequential processes},
  author={Li, Xinyu and Zhang, Yanyi and Zhang, Jianyu and Zhou, Moliang and Chen, Shuhong and Gu, Yue and Chen, Yueyang and Marsic, Ivan and Farneth, Richard A and Burd, Randall S},
  journal={ACM Interactive, Mobile, Wearable and Ubiquitous Technologies},
  volume={1},
  number={3},
  pages={1--20},
  year={2017},
  publisher={ACM New York, NY, USA}
}

@article{qu2020lap,
  title={Lap-net: Adaptive features sampling via learning action progression for online action detection},
  author={Qu, Sanqing and Chen, Guang and Xu, Dan and Dong, Jinhu and Lu, Fan and Knoll, Alois},
  journal={arXiv preprint arXiv:2011.07915},
  year={2020}
}

@article{han2017human,
  title={Human action forecasting by learning task grammars},
  author={Han, Tengda and Wang, Jue and Cherian, Anoop and Gould, Stephen},
  journal={arXiv preprint arXiv:1709.06391},
  year={2017}
}

@inproceedings{aksamentov2017deep,
  title={Deep neural networks predict remaining surgery duration from cholecystectomy videos},
  author={Aksamentov, Ivan and Twinanda, Andru Putra and Mutter, Didier and Marescaux, Jacques and Padoy, Nicolas},
  booktitle={Int'l Conference on Medical Image Computing and Computer-Assisted Intervention},
  pages={586--593},
  year={2017},
  organization={Springer}
}

@inproceedings{marafioti2021catanet,
  title={Catanet: Predicting remaining cataract surgery duration},
  author={Marafioti, Andr{\'e}s and Hayoz, Michel and Gallardo, Mathias and M{\'a}rquez Neila, Pablo and Wolf, Sebastian and Zinkernagel, Martin and Sznitman, Raphael},
  booktitle={Int'l Conference on Medical Image Computing and Computer-Assisted Intervention},
  pages={426--435},
  year={2021},
  organization={Springer}
}

@article{twinanda2018rsdnet,
  title={{RSDNet}: Learning to predict remaining surgery duration from laparoscopic videos without manual annotations},
  author={Twinanda, Andru Putra and Yengera, Gaurav and Mutter, Didier and Marescaux, Jacques and Padoy, Nicolas},
  journal={IEEE Transactions on Medical Imaging},
  volume={38},
  number={4},
  pages={1069--1078},
  year={2018},
  publisher={IEEE}
}

@article{wang2023real,
  title={Real-time estimation of the remaining surgery duration for cataract surgery using deep convolutional neural networks and long short-term memory},
  author={Wang, Bowen and Li, Liangzhi and Nakashima, Yuta and Kawasaki, Ryo and Nagahara, Hajime},
  journal={BMC Medical Informatics and Decision Making},
  volume={23},
  number={1},
  pages={80},
  year={2023},
  publisher={Springer}
}

@inproceedings{fu2024mobile,
  title={Mobile ALOHA: Learning bimanual mobile manipulation using low-cost whole-body teleoperation},
  author={Fu, Zipeng and Zhao, Tony Z and Finn, Chelsea},
  booktitle={Int'l Conference on Robot Learning (CoRL)},
  year={2024}
}

@inproceedings{dosovitskiy2020vit,
  title={An Image is Worth 16x16 Words: Transformers for Image Recognition at Scale},
  author={Dosovitskiy, Alexey and Beyer, Lucas and Kolesnikov, Alexander and Weissenborn, Dirk and Zhai, Xiaohua and Unterthiner, Thomas and  Dehghani, Mostafa and Minderer, Matthias and Heigold, Georg and Gelly, Sylvain and Uszkoreit, Jakob and Houlsby, Neil},
  booktitle={Int'l Conference on Learning Representations (ICLR)},
  year={2021}
}

@inproceedings{sandler2018mobilenetv2,
  title={Mobilenetv2: Inverted residuals and linear bottlenecks},
  author={Sandler, Mark and Howard, Andrew and Zhu, Menglong and Zhmoginov, Andrey and Chen, Liang-Chieh},
  booktitle={IEEE/CVF Int'l Conference on Computer Vision and Pattern Recognition},
  pages={4510--4520},
  year={2018}
}

@inproceedings{he2016deep,
  title={Deep residual learning for image recognition},
  author={He, Kaiming and Zhang, Xiangyu and Ren, Shaoqing and Sun, Jian},
  booktitle={IEEE/CVF Int'l Conference on Computer Vision and Pattern Recognition},  
  pages={770--778},
  year={2016}
}

@article{he2015spatial,
  title={Spatial pyramid pooling in deep convolutional networks for visual recognition},
  author={He, Kaiming and Zhang, Xiangyu and Ren, Shaoqing and Sun, Jian},
  journal={IEEE Transactions on Pattern Analysis and Machine Intelligence},
  volume={37},
  number={9},
  pages={1904--1916},
  year={2015},
  publisher={IEEE}
}

@inproceedings{dwibedi2019temporal,
  title={Temporal cycle-consistency learning},
  author={Dwibedi, Debidatta and Aytar, Yusuf and Tompson, Jonathan and Sermanet, Pierre and Zisserman, Andrew},
  booktitle={IEEE/CVF Int'l Conference on Computer Vision and Pattern Recognition},
  pages={1801--1810},
  year={2019}
}

\end{document}